\documentclass{isprs} % isprs class modified 23-04-2019 (Dennis Wittich)
\usepackage{subfigure}
\usepackage{setspace}
\usepackage{geometry} % added 27-02-2014 Markus Englich
\usepackage{epstopdf}
\usepackage[labelsep=period]{caption}  % added 14-04-2016 Markus Englich - Recommendation by Sebastian Brocks
\usepackage[british]{babel} 
\usepackage[hang]{footmisc}
\usepackage{float}
\usepackage{blindtext}
\usepackage[authoryear]{natbib}
\usepackage{multicol}
 \usepackage{url}
 
\usepackage[dvipsnames]{xcolor}
\usepackage{soul}

\usepackage{xcolor}

\begin{document}

\title{Self-supervised learning - A way to minimize time and effort for precision agriculture?
}

% KAO: Remove extra spacing
\author{
Michael L. Marszalek\textsuperscript{1}\thanks{Corresponding author}  , Bertrand Le Saux\textsuperscript{1}, Pierre-Philippe Mathieu \textsuperscript{1}, Artur Nowakowski\textsuperscript{2}, Daniel Springer \textsuperscript{3}}

% KAO: Remove extra newline
\address{
	\textsuperscript{1 }European Space Agency (ESA) $\Phi$-lab, Frascati, Italy - \\
	(Michael.Marszalek, Bertrand.Le.Saux, Pierre.Philippe.Mathieu)@esa.int\\
	\textsuperscript{2 }World Food Programme (WFP), Rome, Italy - artur.nowakowski@wfp.org \\
	\textsuperscript{3 }Institute of Advanced Research in Artificial Intelligence (IARAI), Vienna, Austria - daniel.springer@iarai.ac.at  \\
}

% If the corresponding author is NOT the final author, always add a % space before the subsequent comma, i.e.
% first author name\textsuperscript{a,}\thanks{Corresponding author} , % second author name \textsuperscript{b}, etc.
% thanks to Niclas Borlin 05-05-2016

%\commission{XX, }{YY} %This field is optional. If filled, XX and YY should be replaced by adequate numbers. See https://www2.isprs.org/commissions/
%\workinggroup{XX/YY} %This field is optional.
\icwg{}   %This field is optional.


\abstract{ Machine learning, satellites or local sensors are key factors for a sustainable and resource-saving optimisation of agriculture and proved its values for the management of agricultural land. Up to now, the main focus was on the enlargement of data which were evaluated by means of supervised learning methods. Nevertheless, the need for labels is also a limiting and time-consuming factor, while in contrast, ongoing technological development is already providing an ever-increasing amount of unlabeled data. Self-supervised learning (SSL) could overcome this limitation and incorporate existing unlabeled data. Therefore, a crop type data set was utilized to conduct experiments with SSL and compare it to supervised methods. A unique feature of our data set from 2016 to 2018 was a divergent climatological condition in 2018 that reduced yields and affected the spectral fingerprint of the plants. Our experiments focused on predicting 2018 using SLL without or a few labels to clarify whether new labels should be collected for an unknown year. Despite these challenging conditions, the results showed that SSL contributed to higher accuracies. We believe that the results will encourage further improvements in the field of precision farming, why the SSL framework and data will be published ~\citep{michael}.
 }

\maketitle
\sloppy

 \section{Introduction}\label{sec:intro}
 %% Crops with EO
 %% Aufbau allgemein -> constrastive -> methoden -> precision agriculture target domain

Food sustainability is one of the grand challenges of the next decades, and a rigorous monitoring of the global food system is needed to allocate our resources ~\citep{FANZO-monitoring-food-food-policy21}. In particular, cropland use monitoring is essential, to assess the supply chain, but also evaluate the impact on natural ecosystems and keep track of the related hidden costs and subsidies~\citep{rockstrom-planet-food-nature20}. Earth Observation (EO) is a global recurring proxy for land use monitoring, which is why it is widely used especially in areas where accessibility and infrastructure are a problem.

Today's opportunity is that there is much more image data available for feature extraction, even if not labeled. In recent years, new methods such as self-supervision have emerged that allow solid representations to be extracted in an unsupervised manner and could provide a more reliable representation for crop type mapping or other precision farming applications. Contrastive learning is a powerful approach to self-supervision, which aims to learn a representation where similar pairs of samples, such as time series of one crop type, are close to each other in the embedding space and different time series are far apart. In addition, only a few available labels do not impose a restriction as with supervised learning. 
Self-Supervised Learning (SSL) was successfully applied to detect changes \citep{lena2021}, using techniques such as pretext tasks or augmentations to learn an invariant representation \citep{GULDENRING2021}. \cite{Baevski22} applied self-supervision for several tasks (NLP, speech, computer vision) with the objective to apply one augmentation suitable for different domains. The core idea was to mask a part of the input instead of using augmentations such as rotation or color distortion which are only suitable for certain use cases. The increasing number of self-supervised learning methods differ mainly in terms of loss function, augmentation or architecture, with the choice of the underlying encoder playing an important role. In this work, we used a transformer (TF) as an encoder, which was verified in previous studies in a supervised manner \citep{breizhcrops2020}. SimCLR was one of the first architectures proposed where augmentation and the use of positive and negative pairs is an important property. Another similar example is MOCO which uses a memory bank in addition to negative and positive pairs. An overview as well as comparison of these siamese networks was presented in ~\citep{chen2020exploring}. We build on the recent SimSiam method, which combines a dual-stream siamese network with various data augmentations on positive pairs of input data ~\citep{chen2020exploring}. SimSiam also achieved promising results with a small batch size and a small number of epochs. 

Self-supervised learning has rarely been used in precision agriculture although it could make an important contribution. For instance, for field-level yield prediction, there are typically very few labels available, which is a limitation for most supervised learning methods. In addition, adapting to a new region or a year with different climatic conditions reduces the effort required to record new labels. In fact, plant morphology varies in different climates, and plant growth and spectral response may change from year to year, due to climatic variations or agricultural practices. Models trained on specific imagery and crop type may therefore not transfer optimally to new regions or future years \citep{belgiu2018a}. From a machine learning perspective, this is framed as domain adaptation. In this context, transfer learning solved similar problems by means of learned knowledge. It used other data sources such as ImageNet to pre-train a neural network and then transfer it to a downstream task \citep{Nowakowski21, Lucas2021}. \cite{Bertrand22} experimented with out-of distribution data and confirmed the power of self-supervised learning with very few labels. ~\cite{Pelletier22} used Thermal Positional Encoding (TPE) for attention-based crop classifiers to classify crop types in different regions in Europe and, most importantly, to reduce the effects of climate on the spectral responses of plants. \cite{Orynbaikyzy2021} identified appropriate Sentinel-1 and Sentinel-2 features to fit a new target region. Reducing the features by eliminating the weather-dependent bands likewise improved the results in our experiments. Another initial application of SSL in the agricultural sector outperformed deep learning approaches with a limited number of labels \citep{GULDENRING2021}. \cite{Agastya2021} successfully applied self-supervised learning for irrigation detection. These promising approaches will save considerable time and reduce the need for labels. 

In this study, SimSiam was applied with and without augmentation. In one experiment, we omitted augmentation and aimed to bring in our existing labels to learn an invariant representation per crop type. Data from previous years already include several variations of time series for each crop type, covering not only small climatological but also soil-related differences. This provided an additional advantage because, unlike normal augmentation with 1D time series, the risk of shifting to another crop type is reduced. \cite{Dwibedi21} followed a similar hypothesis and added nearest neighbors from the data set to find additional positive pairs, assuming that more similar variations would be found this way. 

The objectives of this research were as follows:
\begin{enumerate}
\itemsep=0pt
\item A comparison of supervised learning with self-supervised learning.
\item Experiments with/without augmentation to assess their impact on SSL performance.
\item Analysis to what extent SSL is suitable for the prediction of unknown or deviating years. We assume that this also paves the way for the prediction of crop types in new regions.
\end{enumerate}  

\section{Materials and methods}\label{sec:methods}

\subsection{Study site}
\label{Study site}
 
\begin{figure}[ht!]
\begin{center}
		\includegraphics[width=1.0\columnwidth]{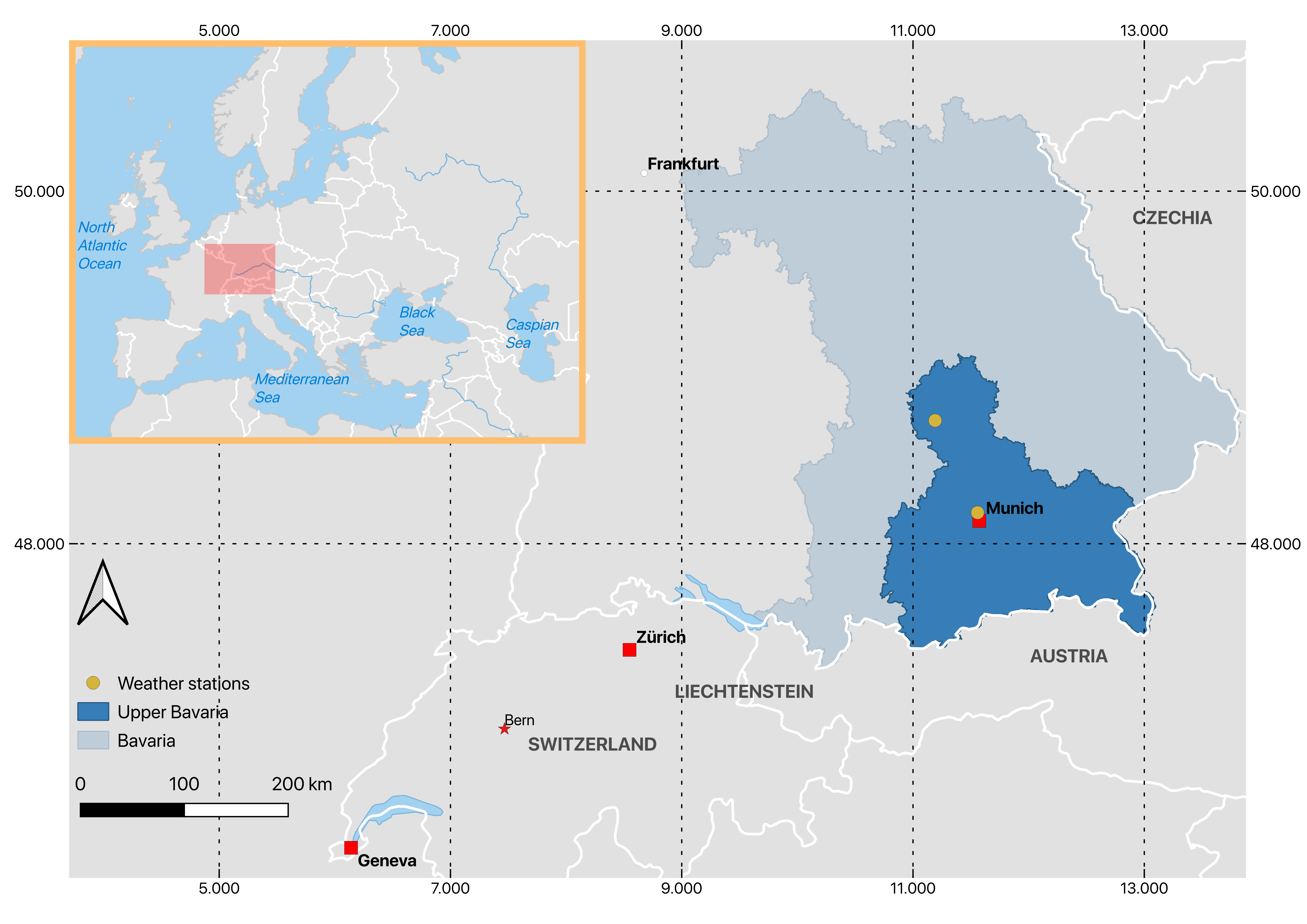}
	\caption{Experimental site in WGS84 (EPSG:4326). All crop type samples were randomly selected in the region of Upper Bavaria. Two weather stations (in orange) were used to generate the climatological overview.}
\label{fig:figure_study}
\end{center}
\end{figure}

The crop mapping task was evaluated on a data set which included the main crop types (corn, winter wheat, winter barley, winter rapeseed, sugar beet, and potato) in Upper Bavaria (Germany), collected for the years 2016, 2017 and 2018 (Figure \ref{fig:figure_study}). It is part of a larger collection, assembled and partly self-created, which included crop types and yields for different regions \citep{michael}. The climatological data in Table \ref{tab:overview} provided better insights into the various climatological conditions.

\begin{table}[h]
\def\arraystretch{1.5}%
	\centering
		\begin{tabular}{|l|c|c|c|}\hline
			  & \textbf{2016}	& \textbf{2017} & \textbf{2018}\\
			  \hline
			 Crop samples& 558 & 600 & 600\\
			 Mean temperature ($^\circ \mathrm{C}$)& 12.3 & 13.0 & 13.7\\
			 Mean precipitation (mm)& 308.5 & 292.9 & 315.4 \\
			 \hline
		\end{tabular}
	\caption{Data overview with average temperature and accumulated precipitation amount per year}
\label{tab:overview}
\end{table}

\begin{figure}[ht!]
\begin{center}
		\includegraphics[width=1.0\columnwidth]{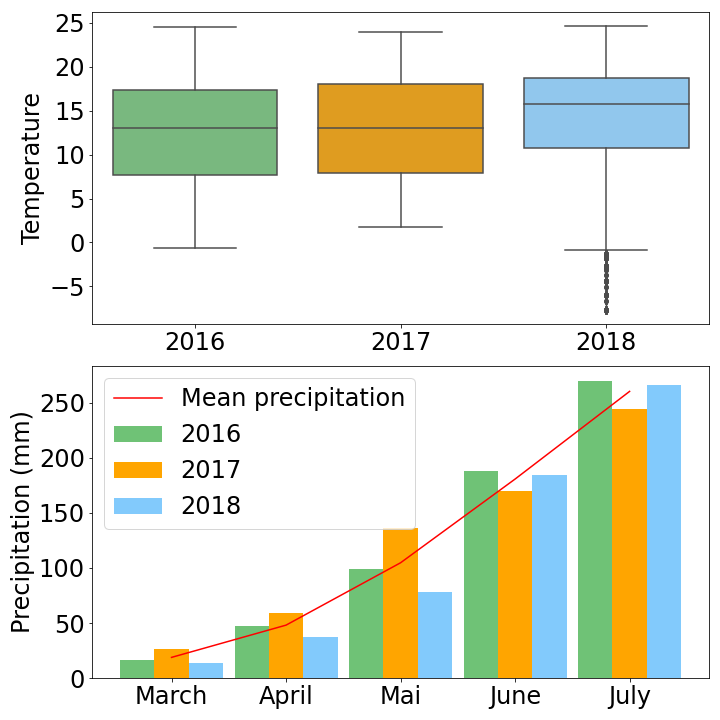}
	\caption{Data on average temperature for each year and an overview of monthly precipitation. Climatological data from March to July were used for the visualisation.}
\label{fig:figure_weather}
\end{center}
\end{figure}
The corresponding climatological overview was visualised in Figure \ref{fig:figure_weather}.
The average temperature was compared for each year and showed that 2018 was 1.4 degrees different from 2016. 2016 and 2017 were more similar in terms of temperature trend and precipitation. Figure \ref{fig:figure_weather} indicates that precipitation in spring 2018 was below normal, which affected crop yields and plant growth. It was visualised as a monthly mean value. Here, climatological data from several fields near two weather stations (highlighted in orange in Fig.~\ref{fig:figure_weather}) were averaged to provide an overview of the period March through July. While this does not reflect the entire region of Upper Bavaria, it is intended to provide an overview of the climatological conditions per year. The crop samples were randomly selected from all over Upper Bavaria.  

\subsection{Sentinel-2}
\label{sentinel}

Sentinel-2 observations were used for the experiments with the self-supervised framework. With a repeat cycle of 2-5 days and its multispectral data in e.g. near infrared (NIR) and short-wave infrared (SWIR), these data allow various applications in agriculture. One drawback is the impact of clouds because, unlike SAR-based missions, clouds limit Sentinel-2's data availability. Nevertheless, the high temporal availability allowed the collection of sufficient data over time in the test region. Initially, all 13 bands were used in this study. A more detailed band description with properties can be found in ~\citep{bands} or ~\citep{Marszalek22}. For this purpose, Level-1C data were downloaded from Google Earth Engine (GEE) ~\citep{gee20}, but not further processed to bottom-of-atmosphere (BOA) reflection information since the focus was the comparison of supervised learning with self-supervised learning. For each field a georeferenced polygon ~\citep{stmelf} was applied for the collection of corresponding field images. The original time series from February to the end of August were interpolated with a resolution of 2 weeks, resulting in 14 time steps for each field. Thus, a sample was represented by 13 band time series with 14 time steps, with each band averaged over the field polygon. For our experiments clouds were partially filtered out, but only in a simplified form, which increased the noise in the time series. GEE provides the ability to filter a scene based on metadata such as "CLOUD\_COVER", but this did not remove the field-level clouds because we used a cloud cover of 20\% for each scene. For 2016, several time series were downloaded incorrectly, which is reflected in the original data set in non-changing values over time. These data were also removed in the pre-processing, which is why slightly fewer samples are available for 2016.

\subsection{Methods}
\label{methods}
Random Forest (RF) \citep{Biau2015ARF} and a transformer (TF) \citep{NIPS2017} were used for the supervised crop type classification task in this study. Overall Accuracy (OA) was used as a measure to evaluate the experiments. 

The semi-supervised crop type classification follows a two-fold training regime. In the initial phase (pre-training) a SimSiam architecture with a transformer as encoder (backbone) is trained to minimise a contrastive loss between augmented sample pairs $x_1$ and $x_2$ of the same crop type and to maximise the loss between pairs of different crop types. The latent representation encoded by the transformer backbone is then used in a supervised classification task (fine-tuning). Here, a linear layer was added as a classification head on the pre-trained backbone. The SimSiam implementation is based on the original work ~\citep{chen2020exploring}. We conducted experiments without and with a few labels of the target year. An important aspect was to verify that the embeddings did not collapse during pre-training. ~\cite{chen2020exploring} suggested that the $l_2$-normalized outputs during training should be close to $1 / \sqrt{dim}$, where $dim$ is the output dimension.

\section{Experiments}\label{sec:results}
 
\subsection{Supervised}
\label{exp1}

The supervised training for RF and the TF was performed in four different scenarios, distinguishable by the data used for training and testing. In the first run the entire data (2016-2018) was used in the train (75\%) and test (25\%) split (E1). The second training was performed with data from 2016 and 2017 only (E2), while data from 2018 was only used in the test set.
The training data of the third (E3) and fourth experiment (E4) is again based on the years 2016 and 2017 but was enriched with 5\% and 10\% of samples from 2018 respectively. 
The first experiment was expected to outperform the other three experiments because it was trained with data from three different years and thus provided the most information about atmospheric and weather patterns. Experiment E2 is the most realistic approach, to illustrate the performance of the framework with respect to unseen data from an entire year.

TF was initialised with $4$ self-attention heads and $3$ layers. All experiments were performed with a learning rate of $0.0016612$, $256$ as batch size and $300$ epochs. As the experiment progressed, bands and time steps were reduced to minimise noise.

\subsection{Self/Semi-supervised}
\label{exp2}

Augmentation is a key factor not only for SSL techniques to learn an invariant representation but also to increase the training data set. 
In the present work we used three augmentations, hereafter referred to as Aug1, Aug2, and Aug3, with Aug1 in particular following a new approach. 

\textbf{Aug1} draws original (non-augmented) pairs of samples from the same crop type and treats them as augmentations of each other. In this way an invariant representation for each crop type is learned. If the pairs are from different years, the model may generalise with respect to different weather patterns. 

\textbf{Aug2} randomly applied noise or drift to the original time series $x_1$ to generate $x_2$. We used a Python framework for time series augmentation ~\citep{tsaug} and parameterised the drift with $max\_drift=0.1$ and $n\_drift\_points=2$. The noise was initialised with $scale=0.02$.

\textbf{Aug3} is a combination of Aug1 with cloud noise simulation. A constant factor (e.g. 7000) was added to all band values for a randomly chosen time step to simulate clouds and learn an invariant representation. Both $x_1$ and $x_2$ were selected from the labeled data set per crop type before random noise was added.

For each augmentation we repeated the same four experiments as discussed in section \ref{exp1}. 
The impact of the different augmentations was analysed in terms of embedded clusters after pre-training and fine-tuning.
The pre-trainig is based on $600$ epochs and a learning rate of $0.0016612$. SimSiam was initialised with 64 features (d\_model), $0.0005$ weight decay and $0.9$ as momentum. The hidden dimension in the projection and prediction head were initialised with $6$ and $14$ as the output dimension respectively. 
The augmentation experiments were performed with 11 and 14 time steps to learn an invariant representation.

\subsubsection{Contrastive}
\label{exp2a}
While the contrastive pre-training aimed to provide a robust feature encoder for the classification task, it is also possible to use this contrastive framework directly as a classification tool. To illustrate the performance of contrastive pre-training, we trained SimSiam according to E1 and E2 and used 2018 data for evaluation. For the classification ($\mathcal{P}$) of a single sample from 2018 ($x_1$) we detected the minimum of the averaged contrastive loss $\mathcal{C}$ between $x_1$ and all $N_c$ samples of each crop type $c$ from 2016 and 2017 ($x_2$) according to 
\begin{eqnarray}
\mathcal{P} = \textrm{min}(\mathcal{L}_c) \ \ \ \ \textrm{with} \ \ \ \mathcal{L}_c = \frac{1}{N_c}\sum_n^{N_c} \mathcal{C}(x_1, x_2^n) \ . \label{eq:contrastive}
\end{eqnarray}

The incorrectly predicted crop types were recorded.

\subsection{Results}

\subsubsection{Supervised}
\label{baseline}

RF and TF were trained with all 13 bands and time steps as a first test. Table \ref{supervised1} summarises the four experiments repeated throughout the paper. %This first test was performed with all data. The second without labels for 2018 followed by 5\%  and 10\%  data for the target year. 
RF improved as the amount of data from the target year increased. In contrast, the transformer had difficulties with this noisy data set which is why the overall accuracy was reduced by about 40\% for 2018 without labels.

\begin{table}[ht!]
\def\arraystretch{1.5}%
	\centering
% 	\resizebox{0.8\columnwidth}{!}{%
		\begin{tabular}{|l|c|c|c|c|}\hline
			  Method & E1 (OA) & E2 (OA) & E3 (OA)  & E4 (OA)  \\
			  \hline
			 RF & 0.92 & 0.80 & 0.84 & 0.87\\
			 TF & 0.93 & 0.54 & 0.86 & 0.62\\
			 \hline
		\end{tabular}
% 		}
	\caption{Supervised benchmark with all bands and time steps: Overall accuracy (OA) with all data (E1); OA with 0\% labels for 2018 (E2); with 5\% labels (E3) and 10\% for 2018 (E4). }
\label{supervised1}
\end{table}

As a next step, bands 1, 2, 3 and 10 were removed because in particular bands 1 and 10 are strongly influenced by aerosols and clouds. Table \ref{supervised2} provides an overview about the results with B4, B5, B6, B7, B8, B8A, B9, B11 and B12. 

\begin{table}[ht!]
\def\arraystretch{1.5}%
	\centering

		\begin{tabular}{|l|c|c|c|c|}\hline
			  Method & E1 (OA) & E2 (OA) & E3 (OA)  & E4 (OA) \\
			  \hline
			 RF & 0.93 & 0.85 & 0.87 & 0.88\\
			 TF  & 0.93 & 0.67 & 0.88 & 0.75\\
			 \hline
		\end{tabular}
	\caption{Supervised benchmark - 14 time steps and 9 bands: Overall accuracy (OA) with all data; OA with 0\% labels for 2018; with 5\% labels and 10\% for 2018. }
\label{supervised2}
\end{table}

Removal of these bands stabilised the prediction of crop types in 2018 for E3 and E4. Tables \ref{supervised1} and \ref{supervised2} show that the results for E4 were worse compared to E3 for RF or TF. The last optimisation step visualised in Table \ref{supervised3} removed the first observations in the time series. An investigation of the time series showed that the signal of the first time steps in 2018 were contaminated by clouds.

\begin{table}[ht!]
\def\arraystretch{1.5}%
	\centering

		\begin{tabular}{|l|c|c|c|c|}\hline
			  Method & E1 (OA) & E2 (OA) & E3 (OA)  & E4 (OA) \\
			  \hline
			 RF & 0.93 & 0.85 & 0.88 & 0.88\\
			 TF  & 0.93 & 0.86 & 0.85 & 0.91\\
			 \hline
		\end{tabular}
	\caption{Supervised benchmark - 11 time steps and 9 bands: Overall accuracy (OA) with all data; OA with 0\% labels for 2018; with 5\% labels and 10\% for 2018. }
\label{supervised3}
\end{table}

Therefore, the time series were shortened by 3 time steps resulting in an observation period from the end of March to the end of August. This significantly affected the accuracy and improved the results for predictions where either 10\% or no samples of 2018 were included in the training. While RF has dominated so far, transformer's predictive capability was emerging here.

\begin{figure*}[ht]
		\includegraphics[width=1\textwidth]{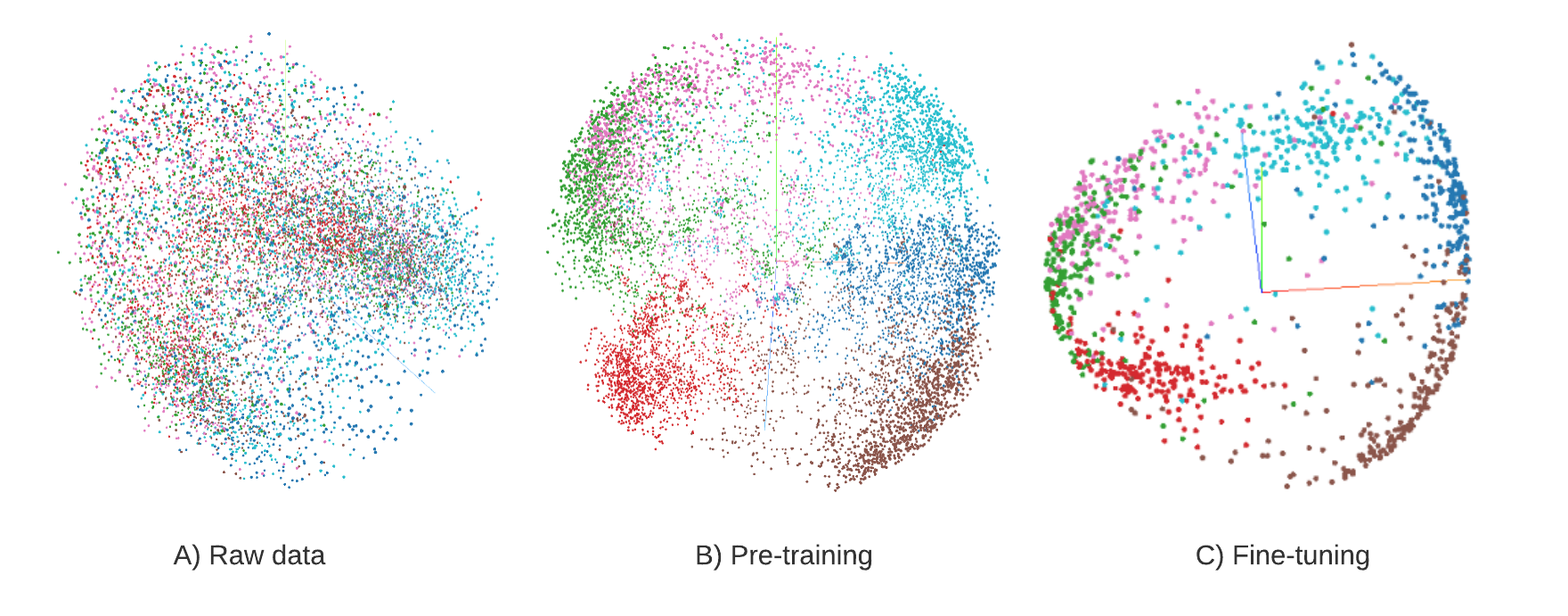}
	\caption{Examples of embeddings generated with principal component analysis (PCA). The six different colors represent the crop types. A) gives an overview of the raw data. B) shows the result after pre-training with SimSiam and Aug1. Here it is already possible to differ between clusters. C) shows the last step and the embeddings after fine-tuning with labels. }
\label{fig:embeddings}
\end{figure*}

\subsubsection{Self/Semi-Supervised}
\label{baseline}

The following results are based on 14 and 11 time steps. As shown in Section \ref{baseline}, the 14 time steps contained cloud noise in the first few weeks, but this is useful for learning a noise-invariant representation. Table \ref{ssl1} provides an overview with Aug1 and Aug2 and 14 time steps. Table \ref{ssl2} focused on Aug1 and Aug3 with 11 time steps and without cloud noise. Comparing the results in Table \ref{supervised2} and Table \ref{ssl1} shows that Aug3 improved the accuracies. The combination of crop invariance and noise in Aug3 also increased the prediction results for E2 to 0.71 OA and E4 to 0.87 OA. For E4, the highest improvement of 16\% was achieved here. Aug1 has only one added value with 10\% samples. The noise-invariant representation also stabilises the results with 10\% no longer collapsing as in Table \ref{supervised2} or Table \ref{supervised1}. Nevertheless, the robustness of RF proved to be evident as well.

\begin{table}[ht!]
\def\arraystretch{1.5}%
	\centering

		\begin{tabular}{|l|c|c|c|c|}\hline
			  14 ts & E1 (OA) & E2 (OA) & E3 (OA)  & E4 (OA)\\
			  \hline
			 Aug1 & 0.92 & 0.54 & 0.68 & \textbf{0.85}\\
			 Aug3 & 0.93 & \textbf{0.71} & 0.87 & \textbf{0.87}\\
			 \hline
		\end{tabular}
	\caption{SSL with 14 time steps: Overall accuracy (OA) with all data (E1); OA with 0\% labels for 2018 (E2); with 5\% labels (E3) and 10\% for 2018 (E4). }
\label{ssl1}
\end{table}

In contrast, Table \ref{ssl2} addressed the results with 11 time steps for a growing season from the end of March to the end of August. Thus, noise caused by clouds was removed in the first 3 time steps in 2018 and changed the presentation of the results. Both, Aug1 and Aug2 performed well in this scenario. Especially the test with 5\% labels reached an accuracy of 0.9 OA compared to 0.85 OA supervised.

\begin{table}[ht!]
\def\arraystretch{1.5}%
	\centering

		\begin{tabular}{|l|c|c|c|c|}\hline
			  11 ts & E1 (OA) & E2 (OA) & E3 (OA)  & E4 (OA)\\
			  \hline
			 Aug1 & 0.90 & 0.85 & \textbf{0.89} & 0.89\\
			 Aug2 & 0.93 & 0.81 & \textbf{0.90} & 0.84\\
			 \hline
		\end{tabular}
	\caption{SSL with 11 time steps:  Overall accuracy (OA) with all data; OA with 0\% labels for 2018; with 5\% labels and 10\% for 2018. }
\label{ssl2}
\end{table}

\begin{figure}[ht!]
\begin{center}
		\includegraphics[width=1.0\columnwidth]{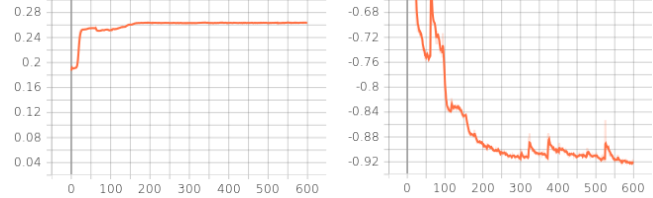}
	\caption{Standard deviation of the $l_2$-normalized output and training loss.}
\label{collapse}
\end{center}
\end{figure}

One observation with Aug1 was the lower accuracy using all data compared with supervised TF (0.9 vs 0.93 OA). Looking at the embeddings visualised with PCA, it is evident that winter wheat and winter barley were difficult to separate. While fine-tuning thereafter separated the two classes well, supervised learning was able to distinguish these two classes more clearly. Figure \ref{fig:embeddings} visualises examples with raw data, pre-training and fine-tuning. Figure \ref{collapse} shows the collapse level of embeddings during the training. Our observations confirmed that convergence near $1 / \sqrt{dim}$ is important to assess whether the embeddings collapse in the pre-training. Evaluation based on training loss alone was not always sufficient. We sometimes observed a collapse of the embeddings during pre-training, which prevented fine-tuning from improving accuracy. Therefore, it was important to observe the collapse level from 300 epochs.

% \subsubsusection{Contrastive}
% \label{baseline}
\textbf{Intermediate evaluation of the backbone} \\
The direct evaluation of the SimSiam network without using a dedicated classification layer, can be understood as an intermediate evaluation of the quality of the latent representation. Table \ref{cont1} shows the crop-wise prediction accuracy of the SimSiam network when evaluated according to equation (\ref{eq:contrastive}). 

\begin{table}[ht!]
\def\arraystretch{1.5}%
	\centering
% 	\resizebox{1.0\columnwidth}{!}{%
		\begin{tabular}{|l|c|c|c|c|c|c|}\hline
			  11 steps & $c_1$ & $c_2$ & $c_3$ & $c_4$ & $c_5$ & $c_6$\\
			  \hline
			 E1 & 0.94 & 0.84 & 0.97 & 0.97 & 0.96 & 1.00 \\
			 E2 & 0.94 & 0.53 & 0.85 & 0.64 & 0.85 & 0.72 \\
			 %E3 & 0.95 & 0.65 & 0.86 & 0.85 & 0.86 & 0.87 \\
			 %E4 & 0.95 & 0.00 & 0.89 & 0.00 & 0.96 & 0.97 \\
			 \hline
		\end{tabular}
% 		}
	\caption{Crop type resolved accuracy of contrastive classification based on 11 time steps for E1 and E2.}
\label{cont1}
\end{table}

In both E1 and E2, we observed a decrease in accuracy for crop type 2 (winter barley), which was more pronounced in E2. Since the evaluation for E1 is based on data that included samples in training, a decline in a single crop type may indicate a systematic problem either in the framework or in the data themselves. 
% The poor performance of the model for both, E1 and E2 for crop type 2 is evident. % Although the evaluation for E1 is based on data that includes samples that were already used in training, the network fails to predict crop type 2 correctly. For E2, where the evaluation is performed on unseen samples from 2018, we observe an overall loss of accuracy and unsurprisingly poor performance for crop type 2.
It is hence of interest to resolve the actual predictions when the network is confronted with samples of crop type 2 from the year 2018 (Table \ref{cont2}). 

\begin{table}[ht!]
\def\arraystretch{1.5}%
	\centering
% 	\resizebox{1.0\columnwidth}{!}{%
		\begin{tabular}{|l|c|c|c|c|c|c|}\hline
			  $\mathcal{P}(c_2)$ & $c_1$ & $c_2$ & $c_3$ & $c_3$ & $c_5$ & $c_6$\\
			  \hline
			 E1 & 0 & \textbf{84} & 1 & 5 & 10 & 0 \\
			 E2 & 1 & \textbf{53} & 2 & 3 & 41 & 0 \\
			 \hline
		\end{tabular}
% 		}
	\caption{Predictions for all $100$ crop type 2 samples from 2018 (correct predictions in bold numbers). }
\label{cont2}
\end{table}
We have noticed a systematic tendency for crop type 2 to be incorrectly identified as crop type 5 (winter wheat). This suggested that crop type 2 and 5 cannot be sufficiently separated by the contrastive framework.

\section{Discussion \& Conclusion}\label{sec:conclusions}
% (research question, relevance, proposed solution, experimental evaluation).

Methods in the field of supervised learning enabled several applications for precision farming \citep{yang, waldner2019a, Vittorio19}. However, supervised learning is in general associated with a large effort since labeled data are required. %In agriculture in particular, this involves a large expenditure of time and effort, since the relevant labels needs to be collected in field trials. 
So far, few experiments were performed with SSL for precision agriculture. In our work, we explored the potential to use SSL for domain adaptation in crop classification.

\textbf{Supervised learning} achieved very good results through noise removal when trained on bands B4, B5, B6, B7, B8, B8A, B9, B11 and B12. Excluding samples from the target year, an overall accuracy of 0.86 was achieved. This resulted in a decrease of about 8\% compared to the test with all the data. We hypothesise that the difference is mainly due to climatological variations ~\citep{MARSZALEK2022}. The transformer had difficulties with noise caused by clouds and improved significantly with the reduction of bands and noisy time steps. RF showed robust performance in all tests regardless of noise.

\textbf{Contrastive pre-training} was used to learn robust representations for each crop type. Augmentation is an important part of SSL and crucial for performance. The introduction of 5\% samples from the target year improved accuracies and outperformed results using supervised learning. While previous attempts were mainly based on typical augmentations such as  rotation or jitter \citep{GULDENRING2021,chen2020exploring}, we took a different approach with Aug1. Learning a crop-invariant representation separated most crop types in the embedding space. Only winter wheat and winter barley were not separable. By directly evaluating the backbone, we showed that these two crop types could not be well distinguished by the contrastive embeddings.
Therefore, we hypothesise that this is the reason why pre-training combined with fine-tuning led to slightly worse results than supervised training in some cases. With 10\% samples we recommend to proceed supervised because here already OA values of around 0.9 were achieved. Aug3 has proven itself in the test with the noise caused by clouds. Here, a significant increase in accuracy was achieved. Successful denoising has also been confirmed in previous studies \citep{Dalsasso_2021}.
The choice of augmentation provides several opportunities to learn a better representation that could also be invariant to regional differences. For example, \cite{Nyborg21} introduced the temporal shift of crop time series in different regions which inspired us to use a drift (Aug2) as augmentation. However, this was only partially successful in learning a shift-invariant representation. Further experiments will be necessary to find a domain-invariant representation for crop types. We also plan to further develop SimSiam based on the lessons learned to more efficiently incorporate unlabeled data into training, which could be important for domain adaption without labels. \cite{Dwibedi21} and \cite{zell22} confirmed that incorporation of unlabeled data improved SSL.

The \textbf{suitability of SSL} was considered for crop classification in this study. We showed that it is essential to introduce few labels for a new year. Although the prediction of crop types in a deviating year was possible, we would like to emphasise that the collection of few samples improved the accuracy and reduced uncertainties. The study targeted crop classification, but we also see potential for other applications in precision agriculture (e.g. yield prediction). We showed that the pre-training improved the results and that a costly collection of labels for a region with already existing data from previous years is only partly necessary. Originally, we assumed that the influence of climatological conditions would be much more significant. Nevertheless, supervised learning and SSL provided good results. Although e.g. a pronounced drought could strongly influence the spectral response of plants, one can question the influence of climate for the classification of crops in this study. The advantage of this test was that we used data from one region and thus minimised the influence of soil and fertilisation. We assumed that farming practices were similar over the years. However, for a change of the target region, these two factors need to be taken into account as they are likely to have an impact on the spectral response. While this study looked at domain adaption, other studies have also identified the benefits for agriculture \citep{GULDENRING2021,Agastya2021}. In general, this study can be an argument for exploring other applications of SSL where data availability is a limitation.

\section*{ACKNOWLEDGEMENT}\label{ACKNOWLEDGEMENTS}
We would like to thank the LIGHTLY team for the support during this research project and the Institute of Advanced Research in Artificial Intelligence (IARAI) for providing the computational resources.

{
	\begin{spacing}{1.17}
		\normalsize
		\bibliography{refs.bib} % Include your own bibliography (*.bib), style is given in isprs.cls
	\end{spacing}
}

\end{document}